\documentclass[journal]{IEEEtran}
\usepackage{amsmath,amsthm}
\usepackage{amssymb}
\usepackage{amsfonts}
\usepackage{graphicx}
\graphicspath{{graphics/}}
\usepackage{tikz,tikz-3dplot}
\usepackage{tikzscale}
\usepackage{xcolor}
\usepackage{bm}
\usepackage{dsfont}
\usepackage{stfloats}
\usepackage[flushleft]{threeparttable}
\usepackage[numbers,sort&compress]{natbib}

\usepackage[switch]{lineno}

\DeclareMathOperator*{\argmin}{arg\,min}

\usepackage[colorlinks]{hyperref}
\hypersetup{colorlinks,breaklinks,linkcolor=blue,urlcolor=blue,anchorcolor=blue,citecolor=blue}
\usepackage{booktabs}
\usepackage{tabularx}
\usepackage{multirow}
\usepackage{lscape}
\usepackage{subcaption}
\usepackage{epstopdf}
\usepackage{booktabs}
\usepackage{caption}
\usepackage{enumitem}

\usepackage{booktabs}
\usepackage{multirow}
\usepackage{array}

\usepackage{float}
\usepackage{microtype}
\usepackage{lineno}
\usepackage{placeins}
\usepackage{xcolor}
\usepackage{algorithm}
\usepackage{algorithmic}

\usepackage{amsmath}
\usepackage{threeparttable}
\usepackage{dcolumn}
\newcolumntype{d}[1]{D{.}{.}{#1}}
\newcommand{\tensor}[1]{\boldsymbol{\mathcal{#1}}}
\newcommand{\mat}[1]{\boldsymbol{#1}}

\ifCLASSINFOpdf
\else
\fi

\usepackage{algorithmic}
\usepackage{algorithm}
\usepackage{xcolor}
\definecolor{darkblue}{rgb}{0.0,0.5,0.5}

\begin{document}
% \linenumbers

% \title{Bare Demo of IEEEtran.cls for\\ IEEE \textsc{Transactions on Magnetics}}
\title{Low-Rank Hankel Tensor Completion for Traffic Speed Estimation}


\author{\IEEEauthorblockN{Xudong Wang, Yuankai Wu, Dingyi Zhuang and Lijun Sun}
% James Kirk\IEEEauthorrefmark{3},
% Montgomery Scott\IEEEauthorrefmark{3}, and
% Eldon Tyrell\IEEEauthorrefmark{4},~\IEEEmembership{Fellow,~IEEE}}
% \IEEEauthorblockA{\IEEEauthorrefmark{2}Twentieth Century Fox, Springfield, USA}
% \IEEEauthorblockA{\IEEEauthorrefmark{3}Starfleet Academy, San Francisco, CA 96678 USA}
% \IEEEauthorblockA{\IEEEauthorrefmark{4}Tyrell Inc., 123 Replicant Street, Los Angeles, CA 90210 USA}% <-this % stops an unwanted space
% \thanks{Manuscript received December 1, 2012; revised August 26, 2015.
% Corresponding author: M. Shell (email: http://www.michaelshell.org/contact.html).}

\thanks{This research is supported by the the Natural Sciences  and Engineering Research Council (NSERC) of Canada, the Fonds de recherche du Quebec - Nature et technologies (FRQNT), and the Canada Foundation for Innovation (CFI). X. Wang would like to thank FRQNT for providing the B2X Doctoral Scholarship. Y. Wu would like to thank the Institute for Data Valorization (IVADO) for providing postdoctoral fellowship.}

\thanks{The authors are with the Department of Civil Engineering, McGill University, Montreal, Quebec H3A 0C3, Canada.
Corresponding author: L. Sun  (Email: lijun.sun@mcgill.ca)}
}


% \markboth{IEEE transactions of intelligent transportation systems}%
% {Shell \MakeLowercase{\textit{et al.}}: Bare Demo of IEEEtran.cls for IEEE Transactions on Magnetics Journals}




% for Transactions on Magnetics papers, we must declare the abstract and
% index terms PRIOR to the title within the \IEEEtitleabstractindextext
% IEEEtran command as these need to go into the title area created by
\IEEEtitleabstractindextext{%
\begin{abstract}

This paper studies the traffic state estimation (TSE) problem using sparse observations from mobile sensors. Most existing TSE methods either rely on well-defined physical traffic flow models or require large amounts of simulation data as input to train machine learning models. Different from previous studies, we propose a purely data-driven and model-free solution in this paper. We consider the TSE as a spatiotemporal matrix completion/interpolation problem, and apply spatiotemporal delay embedding to transform the original incomplete matrix into a fourth-order Hankel structured tensor. By imposing a low-rank assumption on this tensor structure, we can approximate and characterize both global and local spatiotemporal patterns in a data-driven manner. We use the truncated nuclear norm of a balanced spatiotemporal unfolding---in which each column represents the vectorization of a small patch in the original matrix---to approximate the tensor rank. An efficient solution algorithm based on the Alternating Direction Method of Multipliers (ADMM) is developed for model learning. The proposed framework only involves two hyperparameters, spatial and temporal window lengths, which are easy to set given the degree of data sparsity. We conduct numerical experiments on real-world high-resolution trajectory data, and our results demonstrate the effectiveness and superiority of the proposed model in some challenging scenarios.

\end{abstract}

}

\maketitle

\IEEEdisplaynontitleabstractindextext

% \fi
\IEEEpeerreviewmaketitle

\section{Introduction}
Understanding and modeling how traffic state evolves over space and time is a critical problem in traffic flow research and traffic control and management. For example, ramp metering on the highway requires accurate traffic state information to control the traffic flow and mitigate traffic congestion \cite{papageorgiou2002freeway}. Traffic state information, such as flow, speed and density, is generally collected from stationary/fixed sensors (e.g., loop detector/video camera) and moving sensors (e.g., floating cars). The sensors are usually called Eulerian sensors and Lagrangian sensors, respectively. However, neither can provide a complete picture of how the traffic state evolves in space and time. On the one hand, although Eulerian sensors such as loop detector can generate continuous data over time, it only covers a spatial point (or a small segment) where it is installed. Due to the high installation and maintenance cost, loop detectors are installed sparsely for a large-scale transportation network. On the other hand, Lagrangian sensors---such as floating cars with GPS---can contribute real-time traffic state variables (e.g., speed and spacing) along their trajectories \cite{seo2015estimation}. Assuming access to such information from all vehicles on a road segment, one can easily build the full/complete time-space diagram from the trajectory data and estimate flow, density and speed accordingly. However, in practice, data is only available from a tiny fraction of vehicles (e.g., penetration rate less than 5\%). As a result, the traffic state information collected from sparse Lagrangian sensors will naturally be sparse in space and time \cite{ma2021high,thodi2021incorporating}.

Traffic State Estimation (TSE) refers to the task that estimates the complete spatiotemporal profiles of traffic variables based on the partial and sparse observations collected from sensing systems \cite{wang2005real,seo2017traffic}. There are two types of TSE approaches in the literature: model-based and data-driven methods. The model-based methods follow the macroscopic traffic flow---such as the first-order Lighthill-Whitham-Richards (LWR) model or its extensions \cite{herrera2010incorporation, yuan2012real,duret2017traffic}---to estimate traffic state. This approach combines fundamental diagram (FD) and partial differential equations (PDE) with other techniques (e.g., Kalman filter) to depict the traffic dynamics (see e.g., \cite{wang2005real}). The advantages of the model-based methods are that they can reflect the basic traffic principles and provide insights to explain the estimation process. However, model-based methods highly depend on various theoretical assumptions of traffic physics, which might be too abstract and biased to model the noisy and stochastic real-world data. As a result, model-based methods often require reliable prior information to guide the estimation.

Thanks to the availability of large-scale traffic data and recent advances in machine learning, data-driven methods have become increasingly popular in TSE. The key idea of the data-driven approach is to learn and leverage the spatiotemporal correlation/dependency from the traffic data to achieve the estimation task. For example, the characteristics of traffic speed data can be captured by spatiotemporal kernels and then integrated into learning frameworks such as Gaussian processes \cite{treiber2011reconstructing}, convolutional neural networks \cite{thodi2021incorporating}, and graph neural networks \cite{wu2020inductive}. Essentially, such learning-based and data-driven methods require a large amount of training data to effectively learn and characterize the complex spatiotemporal patterns in traffic data.

To deal with scenarios where data is limited and sparse, some recent studies have also tried to integrate the physics of traffic flow to guide the learning process in data-driven methods to achieve higher accuracy. For example, \citet{yuan2021macroscopic} developed a physics-regularized Gaussian process (PRGP) model, and \citet{huang2020physics} and \citet{shi2021physics,shi2021physicsinformed} developed a physics-informed deep learning (PIDL) model individually. We refer interested readers to \cite{thodi2021incorporating},  \cite{yuan2021macroscopic}, and \cite{shi2021physics} for a summary of existing methods and representative studies. These physics-guided and physics-informed models offer a powerful alternative when the observed data is sparse. In these models, the underlying formulation of traffic flow physics (either with learnable parameters or encoded as a prior) plays a critical role in determining TSE performance. In principle, the exact form of the physics model is assumed to be known in advance and consistent over the spatiotemporal domain of interest. However, these assumptions might be too restrictive in real-world applications, and we cannot guarantee that a reasonable and consistent macroscopic model can be learned from real-world data \cite{yuan2021macroscopic,shi2021physics}.

{In this paper, we propose a data-driven model to solve the high-resolution traffic speed estimation problem. Specially, we use highly sparse speed observations collected from floating car data to achieve this goal. We model the spatiotemporal traffic speed data as a multivariate time series matrix (location$\times$time) and treat the estimation of the incomplete traffic speed as a matrix completion problem. One of the most popular approaches to matrix completion is low-rank models. The idea behind such models is that a small number of latent factors determine the traffic state matrix; namely, the spatiotemporal traffic state matrix has low-rank characteristics \cite{wang2021diagnosing}. A variety of matrix completion models has been proposed; however, few of them can deal with high missing rate scenarios especially continuous column/row missing case.}

Inspired by the Hankel matrix/tensor applied in seismic processing \cite{trickett2013interpolation},  image inpainting \cite{jin2015annihilating,yokota2018missing}, time series forecasting \cite{shi2020block}, and signal processing/reconstruction \cite{sedighin2020matrix}, we propose a new traffic speed estimation model named Spatiotemporal Hankel Low-Rank Tensor Completion (STH-LRTC) to address the TSE problem with a high missing rate. {The fundamental assumption of STH-LRTC is that the traffic state data has low-rank characteristics \cite{wang2021diagnosing,chen2020nonconvex}, and the transformed Hankel tensor also shows smooth manifolds in the low-rank space \cite{yokota2018missing}.} Instead of encoding predefined physical priors, STH-LRTC automatically learns to approximate unknown spatiotemporal dynamics and completes missing values in a purely data-driven manner.

The proposed STH-LRTC model has three steps. First, we transform the original incomplete traffic state matrix into a fourth-order Hankel tensor by performing both spatial and temporal delay embedding (i.e., Hankelization). The principle idea of delay embedding is to recursively augment the multivariate time series by repeating portions of them. By doing so, the local information around missing values, such as adjacent locations/timestamps, can be extended in higher dimensions.{ Then, we complete the fourth-order Hankel tensor using a low-rank tensor completion technique. The sum-of-nuclear norm (SNN) \cite{liu2012tensor} is widely used in approximating tensor rank; however, it shows high computation cost and suboptimal solution for fourth-order tensor \cite{lu2019tensor}. To accelerate the algorithm for the fourth-order tensor and have better completion performance, we unfold the tensor to a balanced spatiotemporal matrix \cite{mu2014square} and apply truncated nuclear norm (TNN) \cite{hu2012fast,zhang2012matrix} to approximate the tensor rank.} Finally, we obtain the estimated matrix by performing inverse Hankelization on the completed tensor. The main contribution of this paper is fourfold:
\begin{itemize}
    \item We model spatiotemporal traffic state data (only one variable--speed) as a matrix and apply spatiotemporal delay embedding (i.e., Hankelization) to transform the traffic state matrix into a fourth-order Hankel tensor structure.{As a result, the spatiotemporal correlation/dependency between observations and missing values is naturally augmented without any other additional constraints.}

    \item We characterize the higher-order correlations/dependency in the traffic data by imposing a low-rank assumption on the spatiotemporal Hankel tensor. Thus, we can achieve TSE by performing low-rank completion on the Hankel tensor, and this approach is purely data-driven.

    \item{We incorporate the spatiotemporal tensor unfolding strategy and truncated nuclear norm to improve the computation efficiency and performance for fourth-order tensor.}

    \item{We conduct a challenging TSE, on a real-world dataset, by estimating all traffic speed trajectories using only 5\% of them.} The results show that the proposed STH-LRTC offers superior performance than baseline models.
\end{itemize}

The remainder of this paper is organized as follows. In Section~\ref{sec:pre}, we introduce the preliminaries of this study. In Section~\ref{sec:method}, we introduce the STH-LRTC model in detail and develop an Alternating Direction Method of Multipliers (ADMM) algorithm for model estimation. In Section~\ref{sec:casestudy}, we present a case study using real-world traffic data to evaluate the performance of STH-LRTC. Section~\ref{sec:conclusion} concludes this study and discusses some directions for future research.

\section{Preliminaries}
\label{sec:pre}
\subsection{Notations}
We follow \cite{kolda2009tensor,liu2012tensor} to define all notations. We use lowercase letters to denote scalars, e.g., $x \in \mathbb{R}$, boldface lowercase letters to denote vectors, e.g., $\boldsymbol{x}\in \mathbb{R}^{N}$, boldface capital letters to denote matrices, e.g., $\boldsymbol{X}\in \mathbb{R}^{N \times T}$, and boldface Euler script letters to denote higher-order tensors, e.g., $\tensor{X} \in \mathbb{R}^{I_1 \times I_2 \times \dots \times I_R}$.{We denote the $(i,j)$th entry of a matrix by $\mat{X}_{i,j}$ and the $(a,b,c,d)$th entry of a fourth-order tensor by $\tensor{X}_{a,b,c,d}$. We use $\mat{X}_{i:i+p-1,:} \in \mathbb{R}^{p\times T}$ and $\mat{X}_{:,j:j+p-1}\in \mathbb{R}^{N \times p}$ to denote the sub-matrix that consists of all columns from $i$th row to $(i+p-1)$th row and all rows from $j$th column to $(j+p-1)$th column, respectively.}

The Frobenius norm of a tensor is defined as $\|\tensor{X}\|_F = \sqrt{\sum_{i_1=1}^{I_1} \dots \sum_{i_R=1}^{I_R} x_{i_1\dots i_R}^2}$. The inner product of two tensors of the same size is $\left<\tensor{X}, \tensor{Y}\right>=\sum_{i_1=1}^{I_1} \dots \sum_{i_R=1}^{I_R} x_{i_1\dots i_R}y_{i_1\dots i_R}$. {The tensor unfolding is the process of reordering the elements of an tensor into a matrix. Take a fourth-order tensor for example, the $k$th-mode ($k=1,\cdots, 4$) unfolding denoted as $\tensor{X}_{(k)}$ is reshape the tensor to a $I_k \times (\prod_{i=1,i\ne k}^4 I_i)$ matrix.}

\subsection{Problem description}

The traffic state variables are collected from floating cars in the spatiotemporal domain $L\times W$, where $L$ is the road length and $W$ is the time window length. Given the pre-defined spatial resolution $l_s$ and temporal resolution $l_t$, we can transform the original trajectories into an incomplete traffic variable matrix $\mat{Y} \in \mathbb{R}^{N \times T}$, where $N = L/l_s$ (spatial dimension) and $T=W/l_t$ (temporal dimension). The value of each entry is the average traffic state variable of that entry.

\begin{figure}[htbp]
    \centering
    \includegraphics{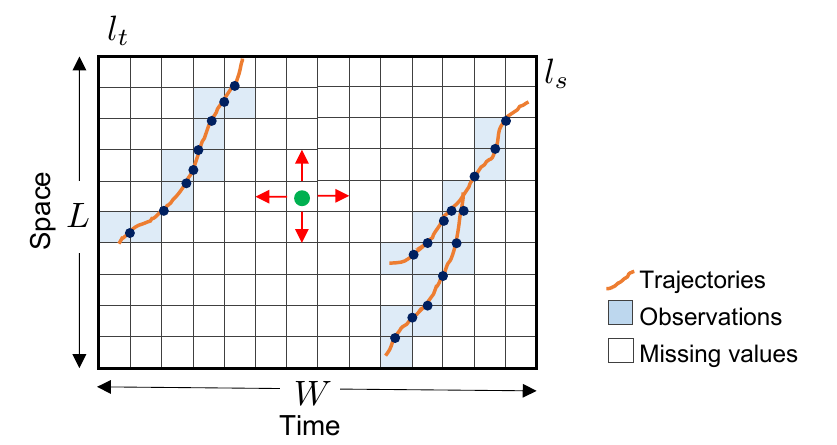}
    \caption{{Matrix representation of traffic state variables collected from floating cars. Traffic state data are discretized and averaged as a matrix. This representation transforms TSE into a matrix completion/interpolation problem. The dark blue dots represent raw spatiotemporal readings from floating cars. The green dot and the red arrows illustrate TV regularization defined in \eqref{eq:TV}.}}
    \label{fig:tse}
\end{figure}

{This study defines TSE as a matrix completion problem using trajectories collected from floating cars. Specifically, we focus on highly sparse matrix completion, e.g., using 5\% trajectories to infer the traffic speed of other unobserved locations or timestamps. In the rest of this paper, we refer to traffic state as the traffic speed variable only, but the model can be expanded to other traffic state variables. Fig.~\ref{fig:tse} shows an example of using three trajectories to estimate the missing traffic speed in both spatial and temporal domains. A few observations lead to a high missing rate ($120/140 \approx 86\% $) and several consecutive periods are entire without observations.}

\subsection{Low-rank-based matrix completion}

{Let $\mat{Z} \in \mathbb{R}^{N \times T}$ be the desired underlying matrix and $\Omega$ be the binary index set to indicate whether the value is observed or not: 0 represents missing value, 1 represents observed value. The low-rank-based matrix completion aims to solve the following optimization problem
\begin{equation}
\begin{aligned}
\label{eq:mf}
  &\min_{\mat{Y}} ~   \text{rank} \left(\mat{Y}\right) + \gamma \mathcal{R}, \quad \text{s.t.}~
  \boldsymbol{Z}_\Omega = {\boldsymbol{Y}}_\Omega, \\
\end{aligned}
\end{equation}
where $\text{rank} \left(\mat{Y}\right)$ denotes the rank of $\mat{Y}$ which can reflect the global consistency, and $\mathcal{R} $ denotes regularization term used to incorporate the spatiotemporal constraint (local consistency) underlying the data to avoid overfitting. The hyper-parameter $\gamma$ is used to trade-off the global consistency and the local consistency.}

% (convex approximation) or nonconvex methods, such as alternating minimization.

{The matrix rank can be approximated by convex approximation, such as matrix nuclear norm. The regularization term is crucial to matrix completion as the correlation between unknown values and observations can be captured. For example, the total variation (TV) regularization is widely used in matrix/tensor completion problem \cite{he2015total}:
\begin{equation}
\label{eq:TV}
\begin{aligned}
        \mathcal{R} =& \sum_{i=1}^{N-1}\sum_{j=1}^{T-1}\{|\mat{Y}_{i,j}-\mat{Y}_{i+1,j}| + |\mat{Y}_{i,j}-\mat{Y}_{i,j+1}| \} \\
        & + \sum_{i=1}^{N-1}|\mat{Y}_{i,T}-\mat{Y}_{i+1,T}| +\sum_{j=1}^{T-1} |\mat{Y}_{N,j}-\mat{Y}_{N,j+1}|.
\end{aligned}
\end{equation}}

{The TV regularization assumes that neighbor locations (same row) and timestamps (same column) are likely to have similar values, e.g., using the known information from the red arrow's direction to estimate the green dot in Fig.~\ref{fig:tse}. However, the performance of matrix completion models will degrade when the consecutive whole rows or/and columns of data are missing as no neighbor information can be learned to estimate the missing values. In this situation, the low-rank structure and the regularization term cannot fully capture the information with a few observations.}

{To tackle the high missing rate with the consecutive missing challenge, we incorporate the spatiotemporal Hankel structure on the matrix and turn the matrix completion problem into a Hankel tensor completion problem. The useful information underlying the data is augmented by the Hankelization process, i.e., enlarging the two-dimensional data to a four-dimensional tensor. The details are described in the following section. }

% As shown in Fig.~\ref{fig:tse}, the trajectories only cover a small portion of entries (10/35) in the spatiotemporal matrix, leading to a high missing rate problem.The objective of TSE is to complete all the unknown traffic variables based on the a few observations, which can be formulated as}

\section{Methodology}
\label{sec:method}

This section introduces a spatiotemporal Hankel low-rank tensor completion (STH-LRTC) model to estimate the traffic variables. At first, introduce a spatiotemporal Hankel tensor transformation in Section~\ref{sec:hankel} and a balanced spatiotemporal unfolding in \ref{sec:squarenorm}. Then, propose an efficient low-rank tensor completion algorithm in Section~\ref{sec:sth}. Finally, give the measurements and convergence criteria in Section ~\ref{sec:implement}. The proposed STH-LRTC model includes three steps: (1) tensor Hankelization, (2) Hankel tensor completion, and (3) inverse tensor Hankelization, which is shown in Fig.~\ref{fig:flowchat}.

\subsection{Spatiotemporal Hankel tensor transformation}
\label{sec:hankel}

% \begin{figure*}[htpb]
%     \centering
%     \includegraphics{Figure/ST_hankel_small.pdf}
%     \caption{Tensor Hankelization and inverse tensor Hankelization illustration when $\tau_s=3,~\tau_t=5$.}
%     \label{fig:hankel}
% \end{figure*}

\begin{figure*}
    \centering
    \includegraphics{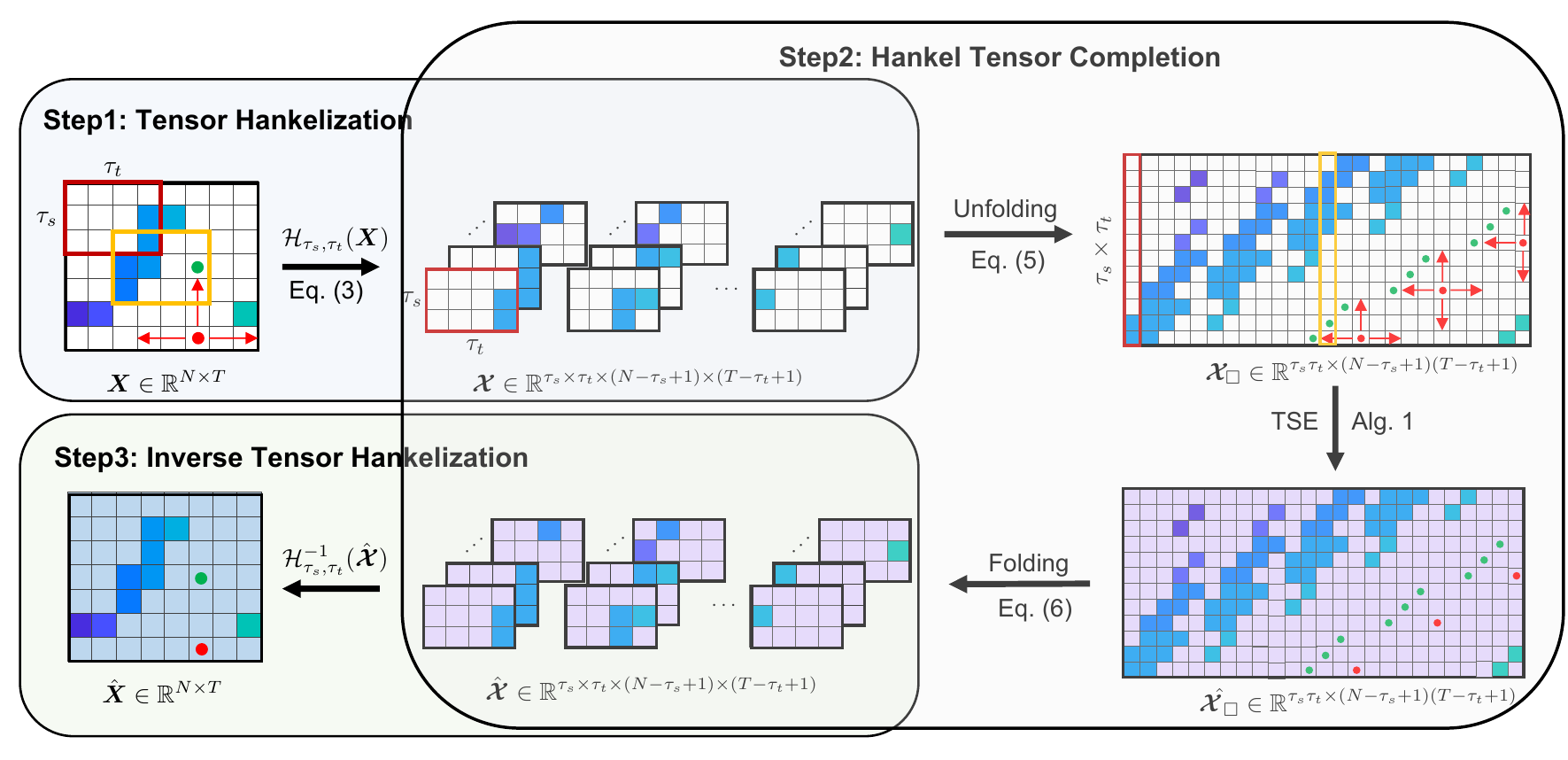}
    \caption{The proposed STH-LRTC includes three steps: (1) Tensor Hankelization (\ref{sec:hankel}), (2) Hankel Tensor Completion (\ref{sec:squarenorm} and \ref{sec:sth}), and (3) Inverse Tensor Hankelization (\ref{sec:hankel}). Top half: the colored entries represent observations, and the white entries represent missing values. Bottom half: the missing values are estimated and colored. The green and red dots with arrows are used as examples to demonstrate the power of Hankelization and spatiotemporal unfolding on low-rank completion. Each $\tau_s \times \tau_t$ patch in the original matrix $\mat{X}$ (e.g., the red and orange rectangles) is vectorized in $\tensor{X}_\square$.}
    \label{fig:flowchat}
\end{figure*}

We define the spatiotemporal Hankelization operation $\mathcal{H}_{{\tau_s,\tau_t}}$ with spatial embedding length $\tau_s$ and temporal embedding length $\tau_t$ as the process to transform a given spatiotemporal matrix $\boldsymbol{X}\in \mathbb{R}^{N\times T}$ to a fourth-order Hankel tensor $\tensor{X}=\mathcal{H}_{{\tau_s,\tau_t}}\left(\boldsymbol{X}\right)\in \mathbb{R}^{\tau_s \times \tau_t \times (N-\tau_s+1) \times (T-\tau_t+1)}$. The spatial and temporal delay embedding operations are introduced simultaneously. Following MATLAB notation, in the Hankel tensor, we have
\begin{equation}
    \tensor{X}_{:,:,i,j} = \boldsymbol{X}_{i:i+\tau_s-1,j:j+\tau_t-1} \in \mathbb{R}^{ \tau_s \times \tau_t},
\end{equation}
for $i=1,\ldots, N-\tau_s+1$ and $j=1,\ldots, T-\tau_t+1$. In general, we expect $\tau_s \ll N$ and $\tau_t \ll T$.

Correspondingly, the inverse Hankelization operation $\mathcal{H}_{{\tau_s,\tau_t}}^{-1}$ is to transform a Hankel tensor $\boldsymbol{\hat{\mathcal{X}}}$ to a matrix $\hat{\boldsymbol{X}}$ \cite{yokota2018missing} by averaging the corresponding entries in the Hankel tensor. For example, for the $(i,j)$th entry in $\hat{\boldsymbol{X}}$, we estimate it by taking the average of all the entries in tensor that are generated from $\boldsymbol{X}_{i,j}$ when performing Hankelization.

An example of the tensor Hankelization and the inverse tensor Hankelization with spatial delay embedding length $\tau_s=3$ and temporal delay embedding length $\tau_t=4$ is shown in Fig.~\ref{fig:flowchat} Step 1 and Step 3, respectively. The delay embedding process can also be extended to multi-dimensional tensors with similar operations \cite{yokota2018missing}.

\subsection{Spatiotemporal tensor unfolding and folding}
\label{sec:squarenorm}

As the underlying true Hankel tensor is low-rank, we develop the following optimization problem to complete missing values:
\begin{equation}
\begin{aligned}
\label{eq:ob_rank}
  &\min_{\tensor{X}} ~   \text{rank} \left(\tensor{X}\right)\\
  & \text{s.t.}~\left\{\begin{array}{l} \tensor{X}=\mathcal{H}_{{\tau_s,\tau_t}}\left(\boldsymbol{Z}\right), \\
  {\boldsymbol{Z}}_\Omega = {\boldsymbol{Y}}_\Omega. \\
  \end{array} \right.
\end{aligned}
\end{equation}

{Computing the tensor rank is an NP-hard problem. One common approach is applying a convex surrogate to approximate tensor rank, such as the sum-of-nuclear norms (SNN) of all tensor unfoldings (matrization): $     \|\tensor{X}\|_*:=\sum_{k=1}^K\alpha_k \|\tensor{X}_{(k)}\|_*$, where $\alpha_k \geq 0$ and $ \sum_{k=1}^K\alpha_k = 1$ \cite{liu2012tensor}. The matrix nuclear norm (NN) is defined as $\|\tensor{X}_{(k)}\|_*=\sum_i \sigma_i(\tensor{X}_{(k)})$, where $\sigma_i(\tensor{X}_{(k)})$ is the $i$th largest singular value of $\tensor{X}_{(k)}$. }

{Another approach is based on heuristic models with nonconvex optimization, such as Tucker decomposition and CANDECOMP/PARAFAC (CP) decomposition. However, such models need prior knowledge to tune the tensor rank, which significantly affects the completion performance. Moreover, \citet{liu2012tensor} show that the convex formulation SNN outperforms the Tucker decomposition and CP decomposition to complete the missing values in third-order tensor scenarios. }

{Although the SNN has been successfully applied in various applications, the computation cost is high when dealing with high-dimensional tensors.} To improve the tensor completion performance, Mu \textit{et al.} \cite{mu2014square} introduce the square norm to approximate tensor rank.{ Instead of calculating the nuclear norm of all tensor unfoldings, the square norm only calculates the nuclear norm of one tensor unfolding -- a more balanced matrix. }

Inspired by this work, we introduce the spatiotemporal unfolding operation that transforms the Hankel tensor $\tensor{X}$ to a matrix $\tensor{X}_\square \in \mathbb{R}^{p\times q }$. Following MATLAB notation, the spatiotemporal unfolding operation can be written as:
\begin{equation}
 \tensor{X}_\square = \text{reshape} \left(  \tensor{X}, \left[p,q\right]\right),
\end{equation}
where $p=\tau_s\times\tau_t$ and $q=(N-\tau_s+1)\times(T-\tau_t+1)$. Correspondingly, the spatiotemporal folding operation that converts the matrix $\tensor{X}_\square$ to the Hankel tensor $\tensor{X}$ can be denoted by the following MATLAB operation:
\begin{equation}
 \tensor{X} = \text{reshape} \left(  \tensor{X}_\square, \left[\tau_s,\tau_t, N-\tau_s+1, T-\tau_t+1\right]\right).
 \label{eq:reshape2}
\end{equation}

{The spatiotemporal tensor unfolding and folding process are shown in Fig.~\ref{fig:flowchat} Step 2. Instead of seeking a more balanced/square matrix in the square norm \cite{mu2014square}, we organize the fourth-order Hankel tensor as a $p \times q$ matrix for two reasons. First, the computation cost of calculating matrix rank is determined by the smaller dimension of the matrix (see variable $\tensor{X}$ update process). Therefore, $\tensor{X}_\square$ is faster than any another reshaped matrices as $\tau_s \ll N,~ \tau_t \ll T$. Second, both $p$ and $q$ consist of spatial and temporal dimensions of the Hankel tensor that can provide the cross information of two domains.}

Combining Hankelization and spatiotemporal unfolding operation provides an effective data-driven and model-free solution to characterize the complex spatiotemporal dependency.{Take the missing value marked by the red dot in Fig.~\ref{fig:flowchat} for example. It is difficult to estimate the red dot as there are no observations at the same column/row with it in $\mat{X}$. However, in $\tensor{X}_\square$, thanks to the Hankelization and spatiotemporal unfolding operation, the red dot is related to the other known values, i.e., the neighbors of the red dot in $\mat{X}$. } The two hyper-parameters $\tau_s$ and $\tau_t$ are easy to set to construct the correlation/dependency structure of the data. When the data becomes highly sparse, one can enlarge the values of $\tau_s$ or/and $\tau_t$ to capture and encode the correlations in an expanded spatiotemporal domain.
% The above transformation is essentially the same as in \cite{jin2015annihilating} for image impainting based on local patches. We can also consider $\tensor{X}_\square$ the block Hankel matrix generated from a 2-d signal \cite{yang1996rank}.

\subsection{Low-rank tensor completion with truncated square norm minimization}
\label{sec:sth}

We use the truncated nuclear norm (TNN) of the spatiotemporal unfolding matrix, i.e., $\|\tensor{X}_\square\|_{r,*}$,  as a more accurate approximation to the rank of tensor $\tensor{X}$  \cite{hu2012fast}. The optimization problem~\eqref{eq:ob_rank} can be transformed into:
\begin{equation}
\begin{aligned}
\label{eq:square}
     &\min_{\tensor{X}}~ \| \tensor{X}_\square\|_{r,*} \\
& \text{s.t.}~\left\{\begin{array}{l} \tensor{X}=\mathcal{H}_{{\tau_s,\tau_t}}\left(\boldsymbol{Z}\right), \\
   {\boldsymbol{Z}}_\Omega = {\boldsymbol{Y}}_\Omega.\\
  \end{array} \right.
\end{aligned}
\end{equation}
where $\|\tensor{X}_\square\|_{r,*} = \sum_{i=r+1}^{\min\{p,q\}}\sigma_i\left(\tensor{X}_\square\right)$ with a positive integer $r < \min\{p, q\}$ \cite{zhang2012matrix}.

{The Alternating Direction Method of Multipliers (ADMM) method is excellent in solving the separable convex optimization \cite{boyd2011distributed}; it has been applied in most existing convex-based matrix/tensor completion models \cite{hu2012fast,zhang2012matrix,mu2014square,liu2012tensor,jin2015annihilating}. We also solve the problem~\eqref{eq:square} by ADMM framework. } The augmented Lagrangian function of \eqref{eq:square} is given to set up an ADMM framework:
\begin{equation}
\begin{aligned}
\mathcal{L}\left(\tensor{X},\boldsymbol{Z},\boldsymbol{E}\right) & =  \| \tensor{X}_\square\|_{r,*} + \frac{\rho}{2}\|{\tensor{X}}-\mathcal{H}_{\tau_s,\tau_t}(\mat{Z})\|_{F}^{2} \\
& + \left\langle\tensor{X}-\mathcal{H}_{\tau_s,\tau_t}(\mat{Z}),\tensor{E}\right\rangle, \\
\end{aligned}
\end{equation}
where $\tensor{E}$ is a dual variable and $\rho > 0$ is a penalty parameter.{The ADMM framework divides the optimization problem \eqref{eq:square} into several suboptimal problems; therefore, the variables $\tensor{X}, \boldsymbol{Z}$, and $\boldsymbol{E}$ are updated alternatively:
\begin{equation}
    \begin{aligned}
        &\tensor{X}^{\ell+1} = \underset{\tensor{X}}{\arg\min~}\mathcal{L}(\tensor{X},\mat{Z}^\ell,\tensor{E}^\ell),\\
        &\mat{Z}^{\ell+1} = \underset{\mat{Z}}{\arg\min~}\mathcal{L}(\tensor{X}^{\ell+1},\mat{Z},\tensor{E}^\ell),\\
        &\tensor{E}^{\ell+1} = \tensor{E}^{\ell} + \rho^{\ell} \left( \tensor{X}^{\ell+1} - \mathcal{H}_{\tau_s,\tau_t}(\boldsymbol{Z}^{\ell+1})\right).
    \end{aligned}
\end{equation}}

The convergence rate becomes slow when applying ADMM to solve many variables and constraints' problems; therefore, we use an adaptive penalty parameter, $\rho^{\ell+1} = \beta \rho^\ell$ with $\beta \in [1.0, 1.2]$ to accelerate the convergence \cite{lin2010augmented}. The inference of updating $\tensor{X}$ and $\mat{Z}$ are shown below.

% \begin{figure*}[!t]
%     \centering
%     \includegraphics{Figure/square.pdf}
%     \caption{An example of unfolding operation that reshape a fourth-order tensor to a balanced matrix and its inverse folding operation when $\tau_s=3$ and $\tau_t=5$.}
%     \label{fig:reshape}
% \end{figure*}

\noindent(A) \textit{Update variable $\tensor{X}$}
\begin{equation}
    \label{eq:admm_x}
    \begin{aligned}
        \tensor{X}^{\ell+1} : &=  \underset{\tensor{X}}{\arg\min~} \frac{1}{\rho} \| \tensor{X}^\ell_\square\|_{r,*} \\
        &+ \frac{1}{2}\|\tensor{X}^\ell- \left( \mathcal{H}_{\tau_s,\tau_t}\left(\mat{Z}^\ell\right) - \frac{1}{\rho^\ell}\tensor{E}^\ell \right)\|_F^2,\\
        &= \text{reshape}\left(\mathcal{D}_{{1}/{\rho^\ell}}\left(\mathcal{H}_{\tau_s,\tau_t} (\mat{Z}^\ell) - \frac{1}{\rho^\ell}\tensor{E}^\ell\right)_{\square}, \text{dim}\right),
    \end{aligned}
\end{equation}
where $\text{dim}=[\tau_s,\tau_t,N-\tau_s+1,T-\tau_t+1]$ and $\mathcal{D}_.(\cdot)$ denotes the singular value shrinkage operator described in the following Lemma 1.

{Lemma 1 \cite{chen2020nonconvex}: Let $\mat{A}=\mat{U\Sigma V}^T$ be the SVD of a matrix $\mat{A}\in\mathbb{R}^{p\times q}$. For any $\tau\geq0$ and $\mat{X}\in\mathbb{R}^{p\times q}$, we have
\begin{equation}
    \mathcal{D}_\tau(\mat{A}) = \argmin \tau\|\mat{X}\|_{r,*}+\frac{1}{2}\|\mat{X-A}\|_F^2,
\end{equation}
where $\mathcal{D}_\tau(\mat{A})=\mat{U}\mat{\Sigma}_\tau\mat{V}^T$ and $\mat{\Sigma}_\tau = \text{diag}(\sigma_1,\cdots,\sigma_r,$ $   \max\{\sigma_{r +1}-\tau,0\}, \cdots, \max\{\sigma_{\min\{p,q\}}-\tau,0\})$. } %\textcolor{red}{randomized svd}

% The solution to this subproblem is
% \begin{equation}
% \label{eq:x}
% \begin{aligned}
%     \tensor{X}^{\ell+1} = \text{reshape}\left(\tensor{X}_\square^{\ell+1}, \text{dim}\right),
% \end{aligned}
% \end{equation}
% where $\tensor{X}_\square^{\ell+1}=\mathcal{D}_{{1}/{\rho^\ell}}\left(\mathcal{H}_{\tau_s,\tau_t}(\boldsymbol{Z}^\ell)_{{\square}}-\frac{1}{\rho}\mathcal{H}_{\tau_s,\tau_t}(\boldsymbol{E}^\ell)_{{\square}}\right)$, $\text{dim}=[\tau_s,\tau_t,N-\tau_s+1,T-\tau_t+1]$, and $\mathcal{D}_{{1}/{\rho^\ell}}(\boldsymbol{M})=\boldsymbol{U}\bm{\Sigma}_{{1}/{\rho^\ell}} \bm{V}^T$ with $\boldsymbol{U}\bm{\Sigma} \bm{V}^T$ being the SVD of $\boldsymbol{M}$. The shrinkage singular value  $\bm{\Sigma}_{{1}/{\rho^\ell}} = \text{diag}(\sigma_1, \cdots, \sigma_r, [\sigma_{r +1}-{1}/{\rho^\ell}]_+, \cdots, [\sigma_{\min\{p,q\}}-{1}/{\rho^\ell}]_+)$, where $[\cdot]_+$ denotes the positive truncation at 0 such that $[\sigma - 1/\rho]_+ = \max\{\sigma - 1/\rho,0\}$.

\vspace{0.5em}
\noindent(B) \textit{Update variable $\boldsymbol{Z}$}
\begin{equation}
    \label{eq:admm_z}
    \begin{aligned}
        \boldsymbol{Z}^{\ell+1} := ~ \underset{\boldsymbol{Z}}{\argmin~}  &\frac{\rho^\ell}{2}\| \tensor{X}^{\ell+1}-\mathcal{H}_{\tau_s,\tau_t}(\boldsymbol{Z}^{\ell})\|_F^2 \\&  + \left<\tensor{X}^{\ell+1}-\mathcal{H}_{\tau_s,\tau_t}(\mat{Z}^{\ell}), \tensor{E}^\ell\right>.
    \end{aligned}
\end{equation}

We denote the objective function in  \eqref{eq:admm_z} by $L_\rho \left(\boldsymbol{Z}\right)$. The optimization can be solved by letting the partial gradient $\displaystyle \frac{\partial L_\rho \left(\boldsymbol{Z}\right)}{\partial \boldsymbol{Z}}=0$. Therefore, the update equation of $\boldsymbol{Z}^{\ell+1}$ is:
\begin{equation}
    \begin{aligned}
    &\boldsymbol{Z}_{\bar{\Omega}}^{\ell+1} = \left(\mathcal{H}_{\tau_s,\tau_t}^{-1}\left(\tensor{X}^{\ell+1} - \frac{1}{\rho^\ell}\tensor{E}^{\ell}\right) \right)_{\bar{\Omega}},\\
    &\boldsymbol{Z}_\Omega^{\ell+1} = \boldsymbol{Y}_\Omega.
    \end{aligned}
\end{equation}

The equations above ensure that only missing values are estimated at each iteration while the observed values remain fixed.

\subsection{Implementation}
\label{sec:implement}
The partially observed speed data from floating cars are used to recover the full traffic speed in the following imputation experiments. Specifically, we randomly select trajectories to generate training index set $\Omega$. We use mean absolute error (MAE) and root mean squared error (RMSE) to measure the performance of the model:
\begin{equation}
\begin{aligned}
    &\text{MAE} = \frac{1}{N_{\text{test}}}\sum_{i=1}^{N_\text{test}} \left|x_i - \hat{x}_i\right|, \\&\text{RMSE} = \sqrt{\frac{1}{N_{\text{test}}}\sum_{i=1}^{N_\text{test}}(x_i-\hat{x}_i)^2},
    \end{aligned}
\end{equation}
 where $N_{\text{test}}$ is the number of test data, $x_i$ is the observation and $\hat{x}_i$ is the estimation.

The following convergence criteria is applied to stop algorithm:
 \begin{equation}
    \frac{||\boldsymbol{Z}^{\ell+1} - \boldsymbol{Z}^{\ell}||_F}{||\boldsymbol{Y}_{\Omega}||_F} < \epsilon,
\end{equation}
where $\boldsymbol{Z}^{\ell+1}$ and $\boldsymbol{Z}^{\ell}$ denote the recovered matrices at two consecutive iterations. Algorithm~\ref{alg:algorithm} summarizes the proposed method STH-LRTC.

\begin{algorithm}[!t]
\textbf{Input}: $\boldsymbol{Y}_\Omega \in \mathbb{R}^{N \times T}$, $\tau_s$, $\tau_t$, $\epsilon$, $\rho$, $\rho_{\max}$, $\beta$ and $r$\\
\textbf{Output}: ${\boldsymbol{Z}}$\\
\textbf{Initialize}: $\ell = 1$
    \begin{algorithmic}[1]
    \STATE $\boldsymbol{Z}^\ell_\Omega = \boldsymbol{Y}_\Omega$ and $\boldsymbol{Z}^\ell_{\bar{\Omega}} = 0$
    \STATE$\boldsymbol{E}^\ell =\boldsymbol{Z}^\ell$
    \WHILE{not converged}
    \STATE $\tensor{Z}^\ell = \mathcal{H}_{\tau_s,\tau_t}(\boldsymbol{Z}^\ell)$ and $\tensor{E}^\ell = \mathcal{H}_{\tau_s,\tau_t}(\boldsymbol{E}^\ell)$
    \STATE $\tensor{X}^{\ell+1} = \text{reshape}\left(\mathcal{D}_{{1}/{\rho^\ell}}\left(\tensor{Z}^\ell_{{\square}}-\frac{1}{\rho^\ell}\tensor{E}^\ell_{{\square}}\right),\text{dim}\right)$
    \STATE $\boldsymbol{Z}^\ell_{\bar{\Omega}} = \left(\mathcal{H}_{\tau_s,\tau_t}^{-1}(\tensor{X}^{\ell+1} - \frac{1}{\rho^\ell}\tensor{E}^{\ell}) \right)_{\bar{\Omega}}$ and $\boldsymbol{Z}^\ell_\Omega = \boldsymbol{Y}_\Omega$
    \STATE $\boldsymbol{E}^{\ell+1} = \boldsymbol{E}^{\ell} + \rho^{\ell} \left( \mathcal{H}_{\tau_s,\tau_t}^{-1}{(\tensor{X}^{\ell+1})} - \boldsymbol{Z}^{\ell+1}\right)$
    \STATE $\rho^{\ell+1} = \min(\beta\rho^\ell, \rho_{\max})$
    %\STATE $e_k = \displaystyle \frac{||\tensor{X}^{k+1} - \tensor{X}^{k}||_F}{||\tensor{X}^{1}||_F}$
    \STATE $\ell = \ell + 1$
    %\IF{$e_k < \epsilon$}
    %\STATE break
    %\ENDIF
    \ENDWHILE
    \end{algorithmic}
    \caption{Spatiotemporal Hankel Low Rank Tensor Completion (STH-LRTC)}
    \label{alg:algorithm}
\end{algorithm}

\section{Case study}
\label{sec:casestudy}
In this section, we apply the proposed STH-LRTC on a real-world traffic dataset to evaluate the TSE performance. Our scenario settings follow the real-world data collection process of floating cars (see e.g., \cite{seo2015estimation}). Our goal is to recover the full traffic speed with high-resolution trajectories observed from a small number of floating cars. We compare the proposed STH-LRTC with several baseline models. {The code for algorithm and experiment is available at github \url{https://github.com/mcgill-smart-transport/traffic_state_estimation}}.

\subsection{NGSIM traffic speed data}

The real-world mobile traffic speed dataset we applied in this paper is an open-source dataset from the Next Generation simulation (NGSIM) program \cite{colyar2007us} on southbound US highway 101 of lane 2. The traffic trajectories are collected by digital video cameras. We select the vehicle trajectories with a global time of fewer than 2700 seconds along a 1500 ft long section. The selected dataset includes 1239 vehicles in total. To organize the trajectories as a matrix, we average the traffic speed with the resolution of $l_s=10$ ft and $l_t=5$ s at first. Then we remove several rows and columns at the beginning/end of the aggregated traffic speed data, which are all-zero values. After pre-processing, the size of the traffic speed matrix is $130 \times 480$, representing a spatiotemporal domain of 1300 ft and 2400 s.

\begin{figure}[htpb]
    \centering
    \includegraphics{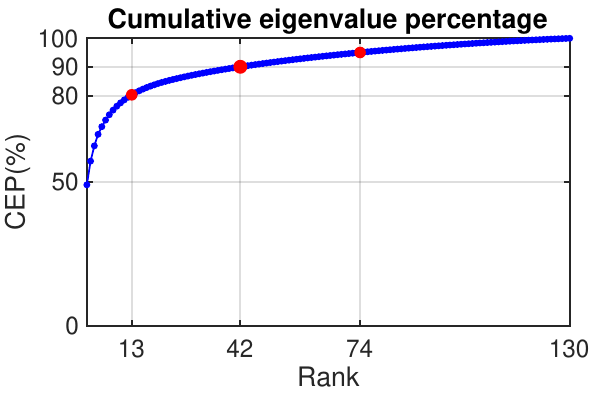}
    \caption{{The cumulative eigenvalue percentage of NGSIM data in which the 80\%, 90\% and 95\% of CEP are obtained by the first 13, 42 and 74 eigenvalues, respectively.}}
    \label{fig:SVD}
\end{figure}

{The proposed model is based on the assumption that the data has low-rank characteristics. Therefore, we use SVD to obtain the cumulative eigenvalue percentage (CEP) of NGSIM data. In Fig.~\ref{fig:SVD}, we can see that a few leading singular values have a significant contribution, e.g., the first 42 singular values cover 90\% of all singular values, showing the low-rank feature of NGSIM data. It has been proved in \cite{yokota2018missing} that the Hankel tensor of a low-rank matrix also shows smooth manifolds in the low-rank space. }
%  Fig.~\ref{fig:NGSIM}(a) shows the  As can be seen, the evolution of traffic state in  NGSIM is more complex than that of the synthetic data, including different lengths of shock waves.

\subsection{Baseline models}
\label{sec:baseline}
{We compare the proposed STH-LRTC method with matrix-based models and Hankel tensor-based models:}
\begin{itemize}
    \item{Matrix factorization with TV regularization (MFTV) defined in \eqref{eq:mf} and \eqref{eq:TV}.}
    \item Adaptive smoothing interpolation method (ASM) \cite{treiber2011reconstructing}: a spatiotemporal kernel weighted method that considers different traffic state (i.e., free-flow/congested) characteristics.
    \item{Spatiotemporal Hankel tensor with Tucker decomposition (STH-Tucker) \cite{yokota2018missing}: using Tucker-based tensor completion model with the spatiotemporal Hankelization to complete the missing values.}
    \item{Spatiotemporal Hankel tensor with sum-of-nuclear-norm (STH-SNN): using SNN \cite{liu2012tensor} to approximate the tensor rank of spatiotemporal Hankel tensor.}
\end{itemize}

    % \item{Bayesian temporal matrix factorization (BTMF) \cite{chen2021bayesian}: a fully Bayesian treatment of matrix factorization by considering temporal dynamics of latent factors.}

{The hyper-parameters in each model greatly affect the imputation performance. For a fair comparison, the baseline models are fine-tuned. We set $\gamma=1$ to balance the low-rank part and the regularization part in MFTV model.} For the ASM model, we set the parameters according to the suggested values in \cite{treiber2011reconstructing} and the characteristics of data: free-flow speed $c_\text{free}=54.6$ ft/s, propagation speed of congestion $c_\text{cong}=-10.0$ ft/s, crossover from congested to free traffic $V_\text{thr}=40.0$ ft/s, and transition width between congested and free traffic $\Delta V = 9.1$ ft/s. The spatial smoothing width $\sigma$ and the temporal smoothing width $\tau$ are 200 ft and 10 s, respectively.

The embedding lengths $\tau_s$ and $\tau_t$ are crucial for the Hankel tensor-based model. On the one hand, a small embedding length cannot fully capture the spatiotemporal information, especially when the missing rate is high. On the other hand, a large embedding length can capture more information leading to heavy computation. In the experiment, we set ${\tau_s}=40$ and  $\tau_t=30$ to obtain the Hankel tensor for STH-Tucker, STH-SNN and STH-LRTC.{We set the core tensor rank $R = [4,4,8,16]$ and $\alpha=[0.1,0.4,0.1,0.4]$ for the STH-Tucker model and the STH-SNN model, respectively. The truncated parameter $r=\lfloor0.05N\rfloor$ \cite{chen2020nonconvex} for the proposed model.}

We set the convergence  $\epsilon=1\times10^{-3}$ for all low-rank based models, the ADMM penalty parameter $\rho=5\times 10^{-6}$ and the adaptive parameter $\beta=1.1$ for all ADMM-based models, i.e., MFTV, STH-SNN and STH-LRTC.

\subsection{Traffic speed estimation results}

Same as the experiment implementation in \cite{thodi2021incorporating}, we randomly select 5\% trajectories/vehicles (62 out of 1239) to construct the incomplete traffic speed matrix.{It has to be noted that the training data is slightly different from the ground truth in the same data entries as we only use the average speed of 5\% vehicles to create the incomplete matrix.} We repeat the experiment 10 times by randomly selecting different training trajectories. The average missing rate is 88\%, i.e., we use 12\% observations to estimate the other 88\% data.

 \begin{table}[htpb]
 \normalsize
\centering
\caption{The average MAE (ft/s) and RMSE (ft/s) with the standard deviation in parentheses in the NGSIM data experiment using 5\% trajectories.}

{\begin{tabular}{@{}llll@{}}
\toprule
\multicolumn{2}{l}{Method}                        & MAE & RMSE \\ \midrule
\multirow{2}{*}{Matrix-based}        & MFTV       & 6.45(0.58)    & 8.79(0.90)     \\
                                     & ASM        & 5.05(0.19)    & 6.62(0.31)      \\\cmidrule(l){1-4}
\multirow{3}{*}{Hankel tensor-based} & STH-Tucker & 7.18(1.01)    & 9.66(1.20)      \\
                                     & STH-SNN    & 6.60(0.83)    & 8.58(1.06)     \\
                                     & Proposed   & \textbf{4.69(0.25)}    & \textbf{6.27(0.40)}     \\ \bottomrule
\end{tabular}}
\label{table:NGSIM}
\end{table}

 Table~\ref{table:NGSIM} shows the average MAE and RMSE obtained from the baseline models and STH-LRTC, numerically demonstrating the missing value completion performance of these models.{Given the highly sparse data, we see that the ASM and the proposed model exhibit better imputation performance than other models. However, the implementation of ASM has a data leakage issue as the hyper-parameters are chosen/estimated based on the ground truth. When ground truth is not accessible/available, the hyper-parameters are very challenging to select and tune, failing to estimate the traffic state.}

{Given the same spatiotemporal Hankel structure, the STH-Tucker and the STH-SNN show larger average MAE and RMSE than the proposed method using truncated square norm. Especially, the STH-Tucker model has the largest average MAE and RMSE with a high standard deviation as it is challenging to set the proper rank for a fourth-order tensor in practice.}

 \begin{figure*}[!htpb]
    \centering
    \includegraphics{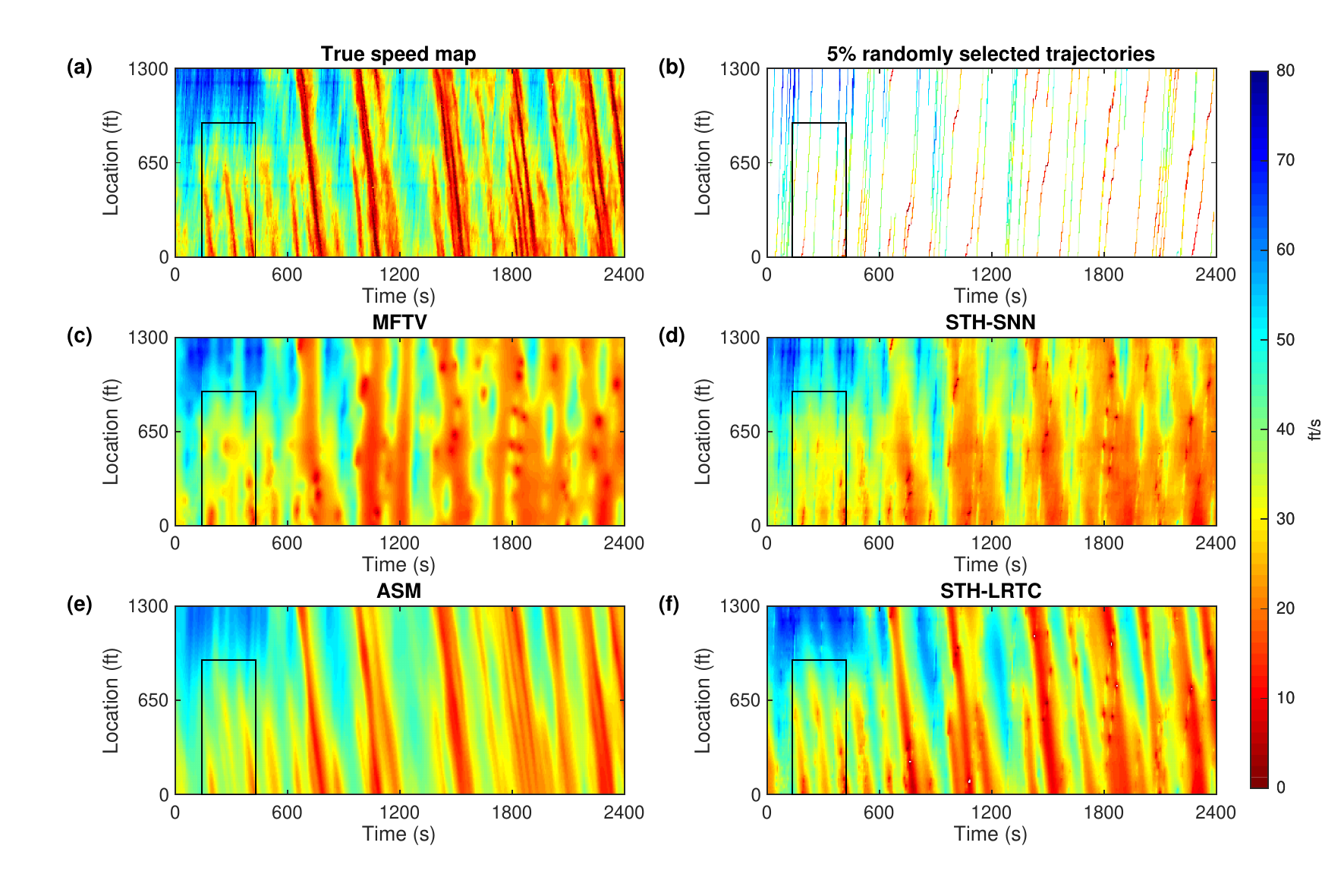}
    \caption{{The NGSIM data experiment: (a) The ground-truth of traffic speed; (b) One random selected 5\% trajectories as training data; (c)The imputation result by MFTV; (d) The imputation result by STH-SNN; (e) The imputation result by ASM. (f) The imputation result by the proposed method STH-LRTC. Three short shockwaves are marked by the black rectangle.} }
    \label{fig:NGSIM}
\end{figure*}

We select one training data and corresponding results except the STH-Tucker model in Fig.~\ref{fig:NGSIM} to better describe the results. The aggregated traffic speed from all the trajectories (ground truth) is shown in Fig.~\ref{fig:NGSIM}(a), in which we can observe several shockwaves backpropagate along the highway. Fig.~\ref{fig:NGSIM}(b) shows one of the selected training data. Several consecutive periods are entirely missing values or have few values, e.g., from 300 s to 340 s and from 1185 s to 1265 s. The high portion of missing grids makes it challenging for models to recover the entire piece of the road traffic speed.

{Fig.~\ref{fig:NGSIM}(c) shows the TSE result of the MFTV model. Due to the TV regularization, close timestamps and locations have similar values. The MFTV model can estimate the missing values but cannot capture details when the missing rate is high. Thanks to the data augmentation power offered by delay embedding, STH-SNN (Fig.~\ref{fig:NGSIM}(d)) shows slightly better performance than MFTV, but it fails to depict different traffic states. }

The completion results of ASM and the proposed model are shown in Fig.~\ref{fig:NGSIM}(e) and Fig.~\ref{fig:NGSIM}(f), respectively. Both models can estimate the shockwaves, but we can see that STH-LRTC shows superior imputation performance. For example, ASM provides over smooth results so that the transitions between waves are blurred, but STH-LRTC can well recover the traffic states even in the transition state. Besides, ASM fails to recover small shock waves (highlight with black rectangular) while STH-LRTC can estimate these small shock waves successfully.

In summary, the proposed STH-LRTC can complete the highly missing value scenario for two reasons. First, the low-rank characteristic can exploit the global information behind the data. Therefore the distinct shapes of stop waves can be discovered instead of being blurred. Second, the delay embedding enhances spatial and temporal information to capture the local consistency in the data to tackle the consecutive missing problem.

{To further explain the superiority of the truncated square norm, we also show the running time of the three Hankel tensor-based models in Table~\ref{table:running} along with MAE and RMSE in Table~\ref{table:NGSIM}. The running time of STH-SNN and STH-LRTC is mainly decided by computing nuclear norm using SVD. The STH-SNN applies unfolding/folding operations and SVD for each iteration four times. However, the STH-LRTC only uses SVD on one balanced unfolding/folding at each iteration. By doing so, the running time significantly decreases, especially when the tensor dimension is high. As for the STH-Tucker model, the running time is mainly caused by core tensor and original tensor size. A small core tensor cannot fully capture the information, but a large core tensor can lead to high computation costs. The slow converge speed is also a problem. }

% \usepackage{booktabs}
\begin{table}[htpb]
 \normalsize
\centering
\caption{The average running time (s) with the standard deviation in parentheses in the NGSIM data experiment using 5\% trajectories.}
{\begin{tabular}{@{}lccc@{}}
\toprule
Method      & STH-Tucker  & STH-SNN    & STH-LRTC  \\ \midrule
Running time & 288.7(88.0) & 290.5(7.8) & 96.3(5.4) \\ \bottomrule
\end{tabular}}
\label{table:running}
\end{table}

%  \begin{figure}[htpb]
%     \centering
%     \includegraphics{Figure/running.pdf}
%     \caption{The running time of three Hankel tensor-based completion model under 10 experiments.}
%     \label{fig:running}
% \end{figure}

\section{Discussion and Conclusion}
\label{sec:conclusion}

This study focuses on recovering the spatiotemporal traffic state data (average traffic speed in our case study) from sparse trajectory-based observations, e.g., a limited number of floating cars. Different from previous TSE studies, which require either physics models of traffic flow or large amounts of simulation data, we propose a purely data-driven and model-free approach based on tensor completion in a spatiotemporal Hankel delay embedded space. In doing so, we treat the spatiotemporal traffic state diagram as a matrix with one dimension representing time and the other representing space. We obtain a fourth-order Hankel tensor structure by applying a two-way delay embedding transform (i.e., spatiotemporal Hankelization). As a powerful data augmentation solution, the spatiotemporal delay embedding not only preserves the global pattern of the original data but also introduces a higher-order dependency/correlation structure within a local spatiotemporal domain.

The proposed framework only involves two hyper-parameters, spatial window length $\tau_s$ and temporal window length $\tau_t$, that affect the domain size of the encoded local correlations. In practice, the two parameters can be determined by the degree of data sparsity and computational cost. We propose an efficient algorithm---STH-LRTC---to complete the Hankel tensor by minimizing the truncated nuclear norm of the spatiotemporal unfolded matrix. We conduct a TSE experiment on real-world high-resolution trajectory data, and our results verify the effectiveness and superiority of the proposed framework, even on an extremely sparse dataset (e.g., 88\% missing values).

There are several directions for future research. %First, this framework can be extended to solve the forecasting task. The current framework is designed for recovering historical traffic state data; however, future traffic states can also be considered missing values in the underlying matrix, and the forecasting problem can then be addressed in the completion framework as well.
First, the transductive nature of matrix/tensor completion requires us to train the model from scratch for each case. The computational cost could be too high for real-time applications when the data is sparse and requires a large embedding $\tau_s\times \tau_t$. In this case, both $p=\tau_s\times \tau_t$ and $q=(N-\tau_s+1)\times (N-\tau_t+1)$ become large in the unfolding matrix but we would still expect $p\ll q$. One potential solution is to applied randomized svd for efficient near-optimal approximation \cite{halko2011finding}. Second, our current method uses the truncated nuclear norm on balanced tensor unfolding to approximate the rank of the Hankel structured tensor; however, this may not be the best approximation given the strong spectral information encoded in the spatiotemporal traffic data. A future research direction is to use alternative rank approximation techniques such as tensor tubal nuclear norm \cite{zhang2016exact,wang2020robust} to estimate the tensor rank.

%{\color{blue} we did not explore the local smoothness. For example, if two patches are close in space and time. They should have }

%\section*{Acknowledgment}

\ifCLASSOPTIONcaptionsoff
  \newpage
\fi



\bibliographystyle{IEEEtranN}
\bibliography{ref}

\begin{IEEEbiography}[{\includegraphics[width=1in,height=1.25in,clip,keepaspectratio]{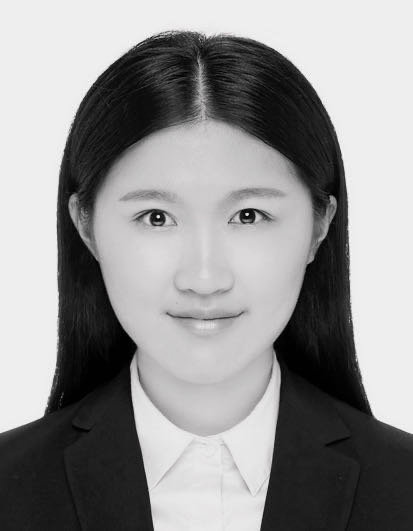}}]{Xudong Wang}
received the B.S. degree in automation from Sichuan University, Sichuan, China, in 2014 and the M.S. degree in automation from Beihang University, Beijing, China, in 2017. She is currently working toward the Ph.D. degree in the Department of Civil Engineering at McGill University, Montreal, QC, Canada. Her research interests include spatio-temporal traffic data mining and anomaly detection.
\end{IEEEbiography}

\begin{IEEEbiography}[{\includegraphics[width=1in,height=1.25in,clip,keepaspectratio]{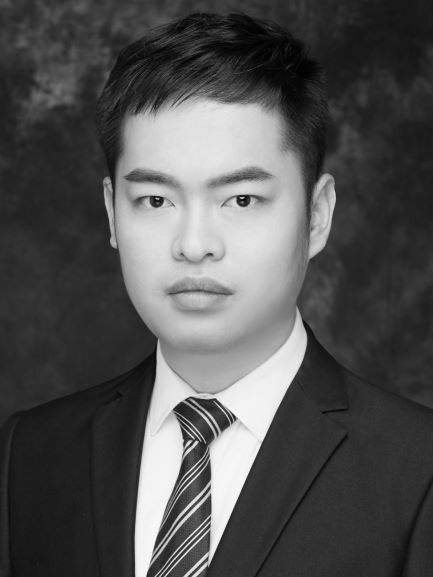}}]{Yuankai Wu}
received the PhD's degree from the School of Mechanical Engineering, Beijing Institute of Technology, Beijing, China, in 2019. He was a visit PhD student with Department of Civil \&
Environmental Engineering, University of Wisconsin-Madison from Nov. 2016 to Nov, 2017. He is a Postdoc researcher with Department of Civil Engineering at McGill University, supported by the Institute For Data Valorization (IVADO). His research interests include intelligent transportation systems, intelligent energy management and machine learning.
\end{IEEEbiography}

\begin{IEEEbiography}[{\includegraphics[width=1in,height=1.25in,clip,keepaspectratio]{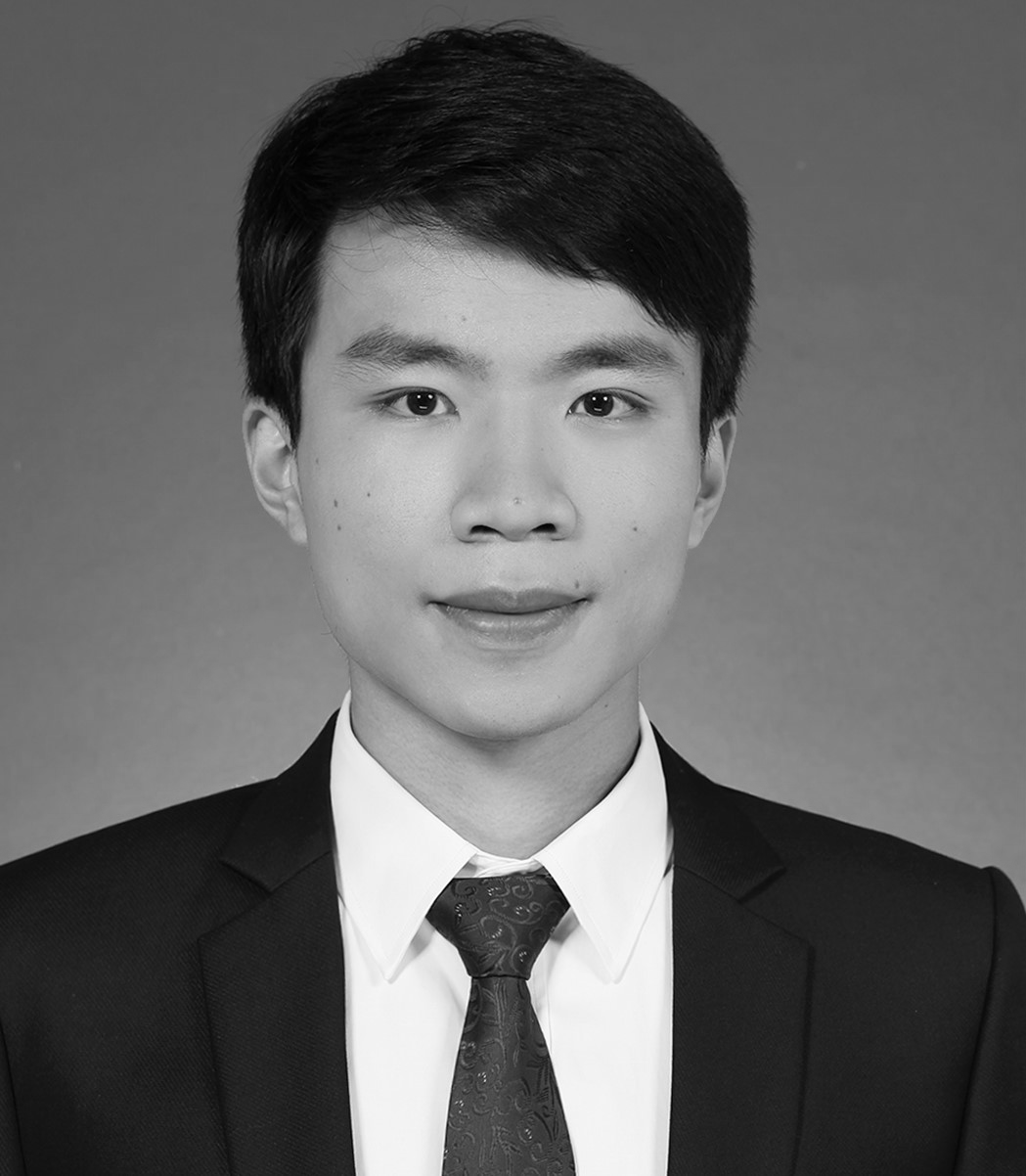}}]{Dingyi Zhuang}
received the B.S. degree in Mechanical Engineering from Shanghai Jiao Tong University, Shanghai, China, in 2019. He was a visit research assistant in the Department of Civil \& Environmental Engineering at National University of Singapore. He is a M.Eng. student in Transportation Engineering expected to graduate in 2021. His research interests lie in intelligent transportation systems, machine learning, and travel behavior modeling.
\end{IEEEbiography}

\begin{IEEEbiography}[{\includegraphics[width=1in,height=1.25in,clip,keepaspectratio]{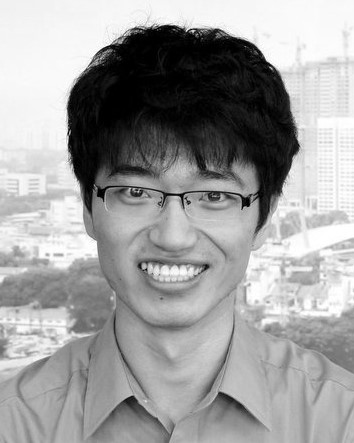}}]{Lijun Sun} (member, IEEE) received the B.S. degree in Civil Engineering from Tsinghua University, Beijing, China, in 2011, and Ph.D. degree in Civil Engineering (Transportation) from National University of Singapore in 2015. He is currently an Assistant
Professor with the Department of Civil Engineering at McGill University, Montreal, QC, Canada. His research centers on intelligent transportation systems, machine learning, spatiotemporal modeling, travel behavior, and agent-based simulation.
\end{IEEEbiography}

\end{document}